\documentclass{vgtc}                          

\graphicspath{{figures/}{pictures/}{images/}{./}} 
\usepackage{amsmath,amssymb}
\usepackage{amsfonts}
\usepackage{times}                     

\usepackage{tabu}                      
\usepackage{booktabs}                  
\usepackage{lipsum}                    
\usepackage{mwe}                       
\usepackage[none]{hyphenat}

\usepackage{comment}

\usepackage{mathptmx}                  

\newcommand{\rev}[1]{\textcolor{black}{#1}}

\onlineid{1149}  

\vgtccategory{Research}

\vgtcinsertpkg

\title{Unwinding Rotations Reduces VR Sickness 
in \\ Nonsimulated Immersive Telepresence}

\teaser{
  \centering
  \includegraphics[width=0.74\linewidth ]{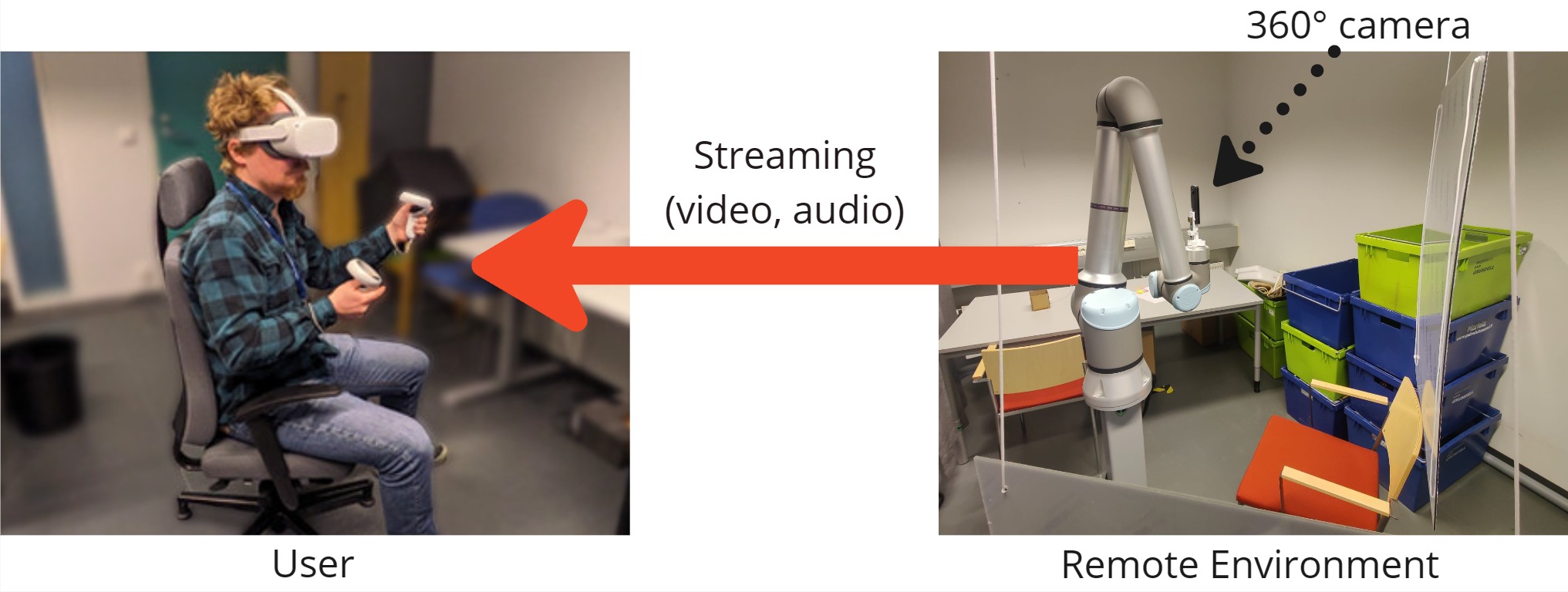}
  \vspace{-1em}
  \caption{Immersive robotic telepresence - a user wearing an HMD embodying a robot arm.}
  \label{fig:setup}
}

\abstract{
Immersive telepresence, when a user views the video stream of a $360^\circ$ 
camera in a remote environment using a Head Mounted Display (HMD), has great potential to improve the sense of being in a remote environment. In most cases of immersive robotic telepresence, the camera is mounted on a mobile robot which increases the portion of the environment that the remote user can explore.
However, robot motions can induce unpleasant symptoms associated with Virtual Reality (VR) sickness, degrading the overall user experience. 
Previous research has shown that unwinding the rotations of the robot, that is, decoupling the rotations that the camera undergoes due to robot motions from what is seen by the user, can increase user comfort and reduce VR sickness. However, that work considered a virtual environment and a simulated robot. 
In this work, to test whether the same hypotheses hold when the video stream from a real camera is used, we carried out a user study $(n=36)$ in which the unwinding rotations method was compared against coupled rotations in a task completed through a panoramic camera mounted on a robotic arm. Furthermore, within an inspection task which involved translations and rotations in three dimensions, we tested whether unwinding the robot rotations impacted the performance of users. 
The results show that the users found the unwinding rotations method to be more comfortable and preferable, and that a reduced level of VR sickness can be achieved without a significant impact on task performance. 
} 
\keywords{Immersive telepresence, Robotic telepresence, Virtual Reality, VR sickness, Cybersickness.}

\usepackage{cite}
\usepackage{algorithmic}
\usepackage{graphicx}
\usepackage{textcomp}
\usepackage{xcolor}

\def\threesixty{$360^\circ$}

\usepackage{graphicx}
\usepackage{subfigure}
\usepackage{multirow}
\usepackage{url}
\usepackage{enumitem}

\usepackage{acronym}
\acrodef{VE}[VE]{\emph{Virtual Environment}}
\acrodef{RE}[RE]{\emph{Remote Environment}}

\author{
Filip Kulisiewicz\thanks{e-mail: 259316@student.pwr.edu.pl}\\
\parbox{1.4in}{\scriptsize \centering  Wrocław University of Science and Technology \\ University of Oulu}
\and
Basak Sakcak\thanks{e-mail: basak.sakcak@maastrichtuniversity.nl}\\
\parbox{1.4in}{\scriptsize \centering Maastricht University \\ University of Oulu }
\and
Evan G. Center\thanks{e-mail: evan.center@oulu.fi}\\
\scriptsize University of Oulu
\and
Juho Kalliokoski\thanks{e-mail: juho.kalliokoski@oulu.fi}\\
\scriptsize University of Oulu
\and
Katherine J. Mimnaugh\thanks{e-mail: kmimnau2@illinois.edu}\\
\parbox{1.4in}{\scriptsize \centering University of Illinois Urbana-Champaign \\ University of Oulu} 
\and
Steven M. LaValle\thanks{e-mail: steven.lavalle@oulu.fi}\\
\scriptsize University of Oulu
\and
Timo Ojala\thanks{e-mail: timo.ojala@oulu.fi}\\
\scriptsize University of Oulu
}

\begin{document}

\maketitle


\section{Introduction}

Immersive robotic telepresence is a way of embodying a physical robot with the use of an HMD.
To effectively leverage the immersive potential of an HMD, a robot is equipped with a 360° camera. Such a combination allows a user to look around freely compared to experiencing the remote location through a view from a non-panoramic camera displayed on a 2-D screen. This wider field of regard can increase immersion, or the extent to which the system can support natural sensorimotor contingencies for perception
\cite{immersion_def}.
Although higher levels of immersion can improve the user experience, the use of an HMD for robotic telepresence can come with the disadvantage of inducing VR sickness in users.  

VR sickness is comprised of a number of uncomfortable sensations that can arise as the result of using an HMD \cite{SSQ20Limit}.
Its deleterious consequences have stimulated research on 
different methods to improve user comfort \cite{Ang_Quarles_2023,Biswas_Mukherjee_Bhattacharya_2024}. 
In the case of immersive robotic telepresence, previous work to decouple the view of the user from the robot rotations has shown promising results in simulation, with an increase in comfort and a decrease in VR sickness when the rotations are unwound \cite{Unwinding, cash2019improving}.
\rev{ 
However, in previous work researchers evaluated the unwinding rotations method in environments built with Unity 3D and only allowed certain camera movements, that is, planar motion and a single degree of rotational freedom (yaw rotation about an axis perpendicular to the ground) \cite{Unwinding} or with two degrees of rotational freedom (yaw rotation and pitch rotation about an axis parallel to the ground) \cite{cash2019improving}. In this paper, we seek to answer the question of whether the unwinding rotations method is effective when VR sickness is induced in a real-world context, thus greatly increasing its generalizability. Our research improves upon previous work in three important ways. First, we consider a nonsimulated environment using pre-recorded videos of an office environment. VR sickness is often attributed to sensory conflicts \cite{Reason_1978}, and humans have built up much stronger sensory expectations to real environments over their lifetimes than to \rev{three-dimensional} rendered environments where things may or may not always closely match their real-world counterparts. Second, we consider rigid body transformations composed of translations along and rotations about three-dimensional vectors.
Finally, we also evaluate whether unwinding rotations can result in an improvement in task performance. } 

The main contribution of this paper is a user study in a nonsimulated environment showing that the unwinding rotations method reduces VR sickness, increases the comfort of an immersive experience, and is preferred by users. 
We performed the user study using a Remote Environment (RE) based on pre-recorded 360$^\circ$~videos taken during robot arm movement around a room for an inspection task. Our findings can improve user experience in real-life robot teleoperation, with particular relevance to environments where \rev{motion} in any direction is possible, such as underwater, in the air, or in outer space. 

\section{Related Work}
Robotic telepresence is a combination of technologies that enables users to navigate in and interact with remote environments as if they were physically present in them. It is achieved by combining robotics with remote communication technologies. Telepresence robots have the potential to provide better access to experiences from any location without having to physically travel. 
They have already been used for applications in various domains, such as education \cite{RobotEducation}, disaster response and rescue \cite{RobotDisasterRescue}, 
and medical teleconsultation \cite{RobotMedicalConsultation}.
These platforms differ drastically from one another in terms of their levels of immersion, 
from a humanoid robot that allows for ``instinctive control of the entire robot body" \cite{RobotToyota} to 
robots that can be thought of as ``embodied video conferencing on wheels" \cite{RobotsUseCases}.
Despite the availability of various robot platforms equipped with video conferencing systems \cite{RobotMobileSocial} and research in that field, complex remote interactions, such as solving a puzzle that requires human communication, performed through currently available systems are still far from the results achieved by face-to-face interactions
\cite{TelepresenceCooperation}.

What can improve the telepresence experience is an increase in the perception of ``being present" in a remote environment by, for example, increasing immersion \cite{Underwater_Immersion}.
An increase in immersion leads to a reduction in mental workload and improved task performance
\cite{PlaceIllusionAndPlausibility_Copresence}. However, this increase also brings in complications.
Telepresence systems with higher levels of immersion
may result in higher levels of VR sickness \cite{Chang_Kim_Yoo_2020,Tian_Lopes_Boulic_2022,CybersicknessFactors}.
We use VR sickness as a name for any unintended and uncomfortable side effects from using a VR system \cite{VistualRealityLaValle}.
Common symptoms of VR sickness include nausea, dizziness, increased salivation, sweating, headache, fatigue, and eye strain \cite{CybersicknessFactors}. In addition to being unpleasant in the moment, the effects of VR sickness can last for some time after the end of the 
experience \cite{Duzmanska,stanney1998aftereffects}. 
VR sickness can also degrade user experience by reducing feelings of presence \cite{Weech} and interfering with attention to and performance on a 
task \cite{attention}. 
Though there are several theories regarding the cause of VR sickness \cite{SSQ20Limit}, one of the most widely discussed is the sensory conflict theory \cite{Reason_1978}.
As a user moves in VR, motion is perceived differently by the eyes that see the movement and by the vestibular organ that does not react in the same way because the user often does not move in the physical world. It is the difference between this pattern of conflicting incoming sensory information and the expectation of typical sensory information, where visual and vestibular information match, which leads to sensory conflict.

An important application domain for immersive telepresence robots is inspection of a remote environment. 
However, sustained use of HMDs for teleoperation in robotic inspection tasks has been associated with increases in VR sickness \cite{Schmidt2014_usability_of_telemanipulation}. Research on underwater teleoperation indicates that high immersion can improve situational awareness and efficiency, but prolonged use can lead to discomfort and usability challenges \cite{HMDs_in_remote_Remotely_Operated_Vechicle, underwater_expeditions}. 
Although HMDs enhance the control and functionality of teleoperated systems by increasing the level of immersion compared to flat screens, the trade-off is the likelihood of experiencing VR sickness, necessitating techniques to reduce this discomfort \cite{underwater_expeditions, Higher_immersion_for_HMD}. 

Previous research has attempted to reduce VR sickness by, for instance, stabilizing camera motions \cite{litleskare2019camera}, dynamically modifying the field of view \cite{fernandes2016combating}, or providing a reference frame \cite{duh2001independent}. Although these methods could provide benefits in addition to the current proposed method, they do not specifically address the significant role of rotational motion in inducing VR sickness, which has been indicated as one of the key factors contributing to an increase in discomfort during immersive virtual experiences \cite{CybersicnessKemeny2020}. 
In particular, rotational motion which is unexpected can result in higher levels of VR sickness \cite{Teixeira_Miellet_Palmisano_2022,Teixeira_Miellet_Palmisano_2024}. 
As a consequence, it is important to consider the experience of rotational motion for users. 

One method that can address these issues is to decouple the viewpoint orientation of the user from the orientation of the robot \cite{Unwinding, cash2019improving}. 
A previous study \cite{joystickUseRotation} explored the potential of physically rotating the head as a rotation method. The researchers found that controlling rotation via head motion while still using a joystick for translation reduced the amount of time needed to complete a navigational task in an HMD when compared to using a joystick for both translation and rotation. This decoupling also provides the benefit that the user knows when rotational motion will occur, thus precluding any VR sickness inducing contribution from rotation which has not been anticipated. 

\begin{figure}
\centering   
\subfigure[Coupled Rotations]{\includegraphics[width=.47\columnwidth]{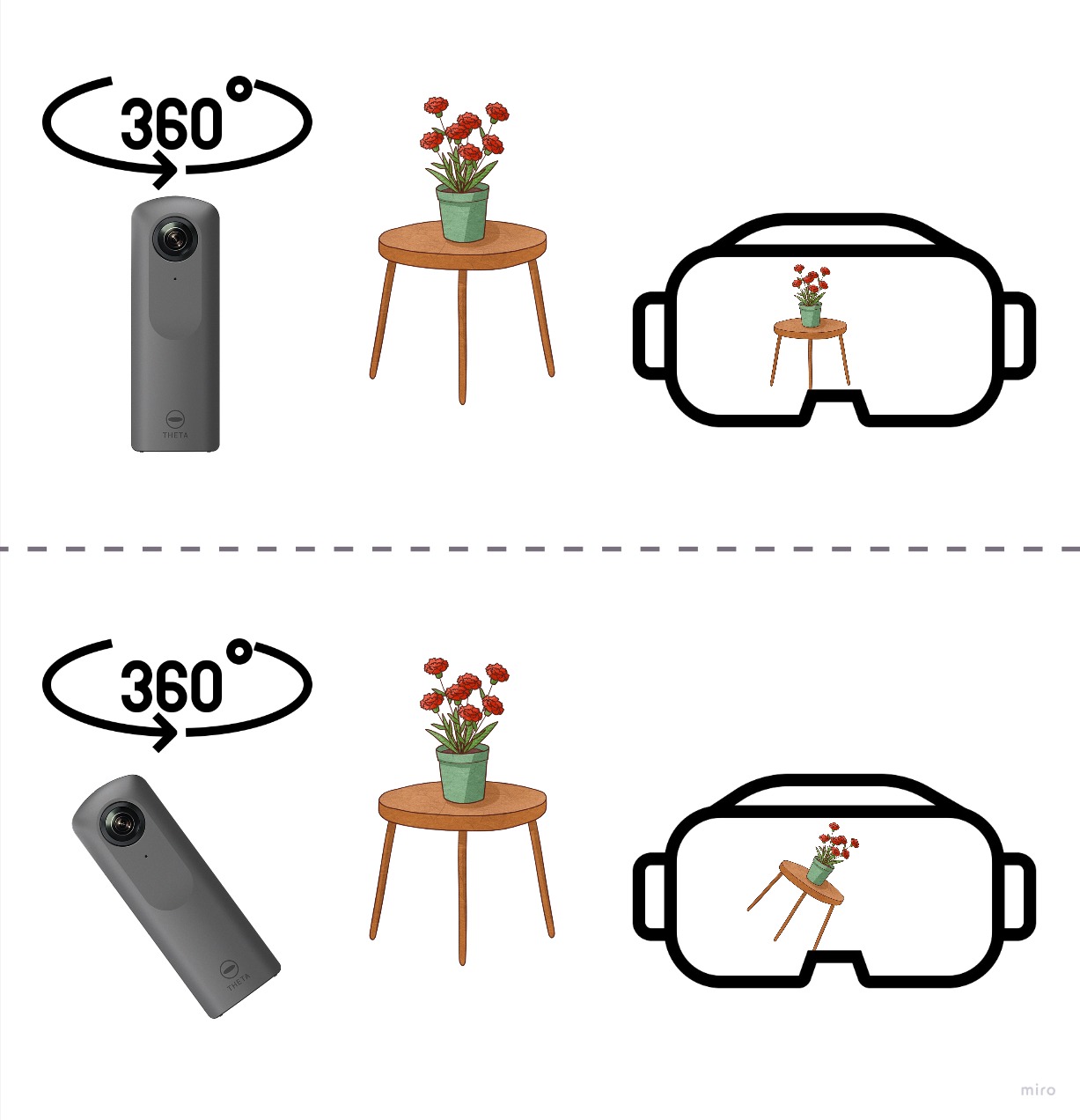}
    \label{fig:unmodViewpoint}
}
\subfigure[Unwinding Rotations]{
    \includegraphics[width=.47\columnwidth]{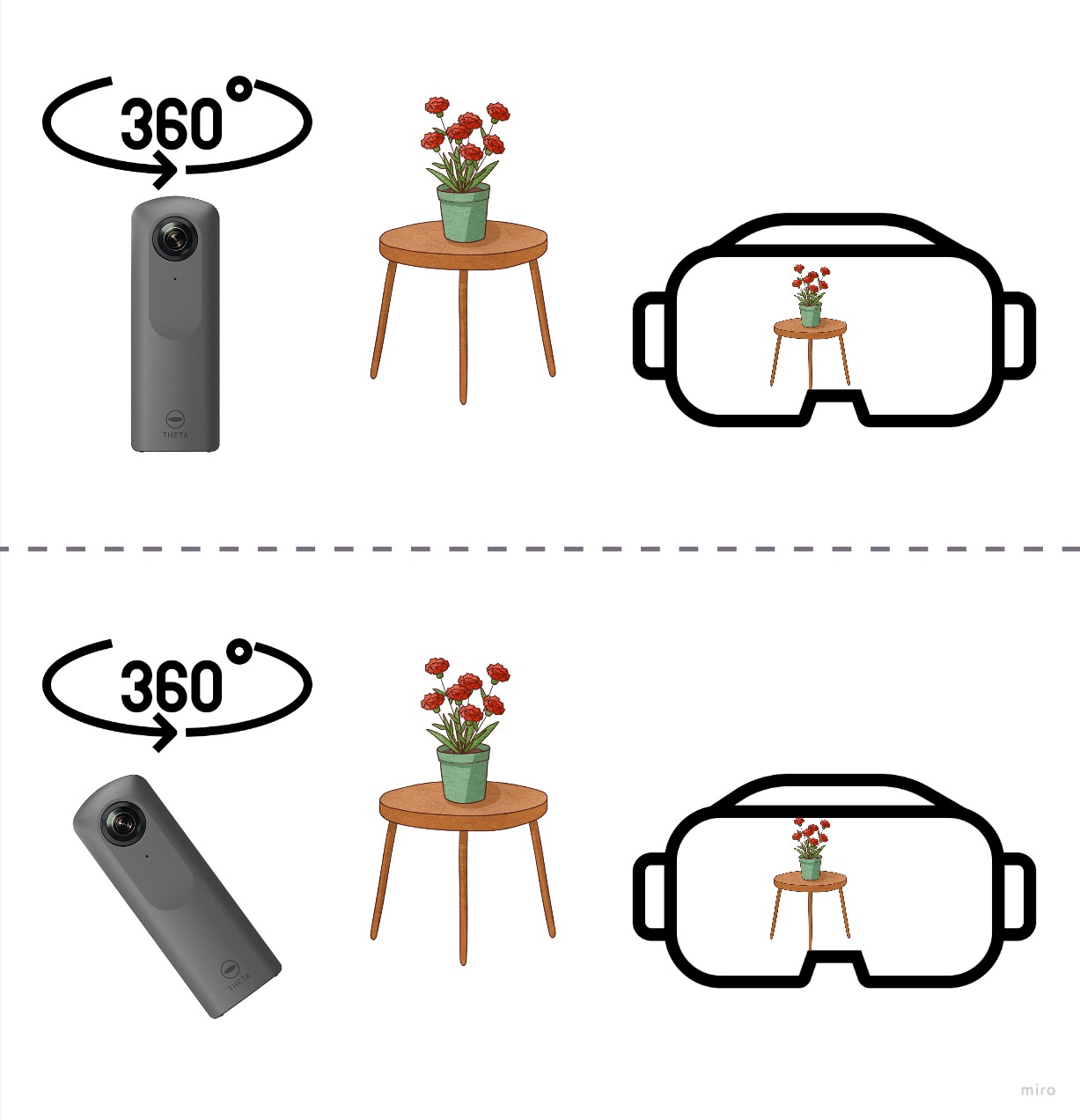}
    \label{fig:stabilViewpoint}
}
\vspace{-2mm}
\caption{(a) \rev{When camera rotations are not unwound, the} user's viewpoint is coupled with the camera rotations. (b) The camera rotations are unwound to decouple the user's viewpoint from the camera rotations.}
\label{fig:abstrViewpoint}
\vspace{-1.5em}
\end{figure}

\section{Unwinding Rotations}\label{sec:UnwindingRotations}
\rev{In this section, we briefly explain how unwinding rotations method works. Suppose the user does not move their head and the camera undergoes a rigid body transformation, that is, it rotates and/or translates in the three-dimensional space. What the user sees will change based on the motion of the camera. When the rotations are unwound, we are in essence canceling out the rotational motion as viewed by the user. Then, even if the camera is moving, if the user does not move their head they will not see any rotation happen. Translational motions of the camera will still be visible from the changing point of view of the camera, but any visible rotational motion will arise from user head motion alone and not the motion of the camera through space. Fig.~\ref{fig:abstrViewpoint} illustrates the change in user's experience of watching the video stream of a \threesixty~camera moving in a remote environment via an HMD when unwinding rotations.}

\begin{figure}
\centering   
\subfigure[]{\label{fig:Cam_base}\includegraphics[width=.33\columnwidth]{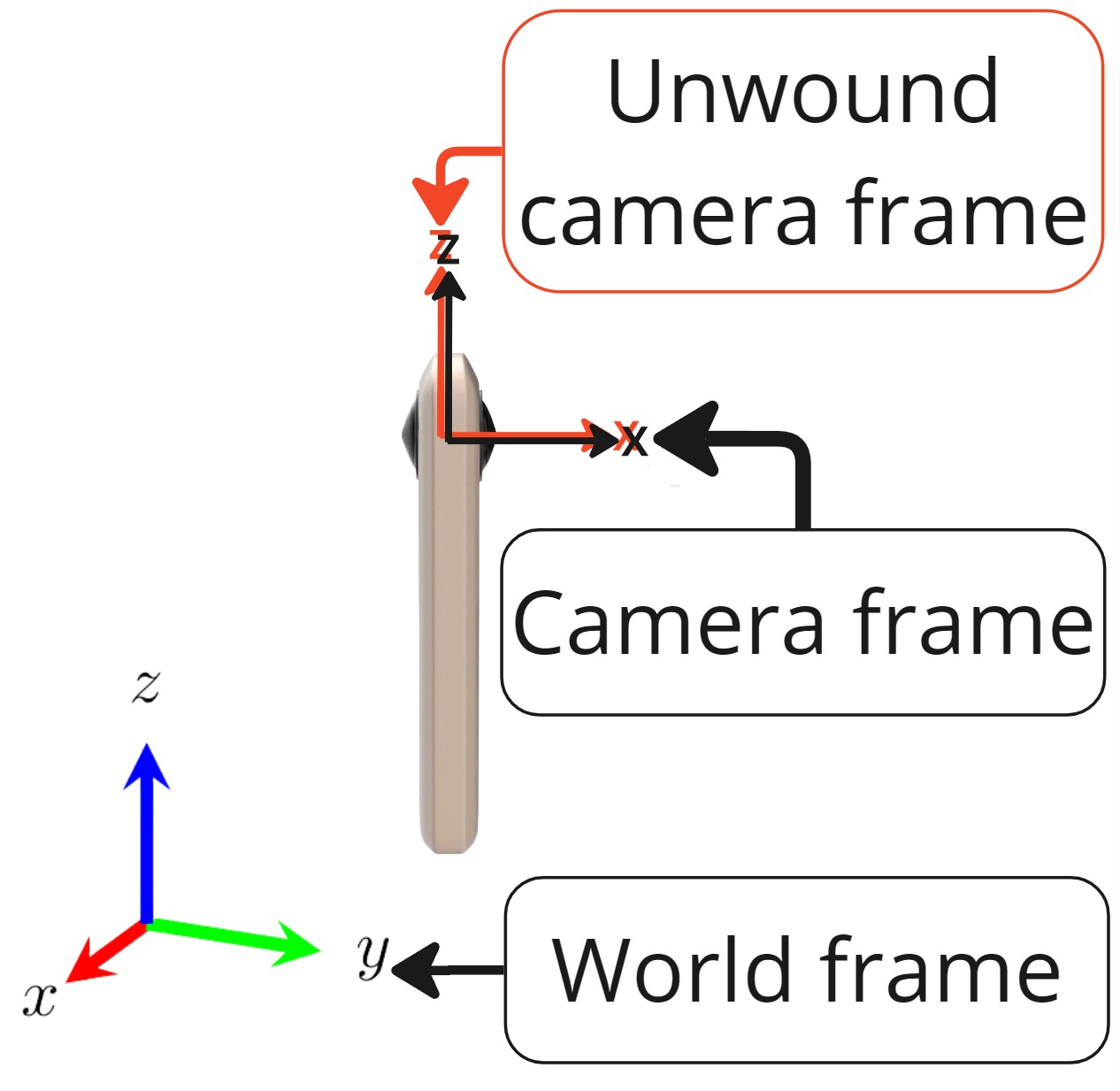}}
\subfigure[]{\label{fig:Cam_rot}\includegraphics[width=.3\columnwidth]{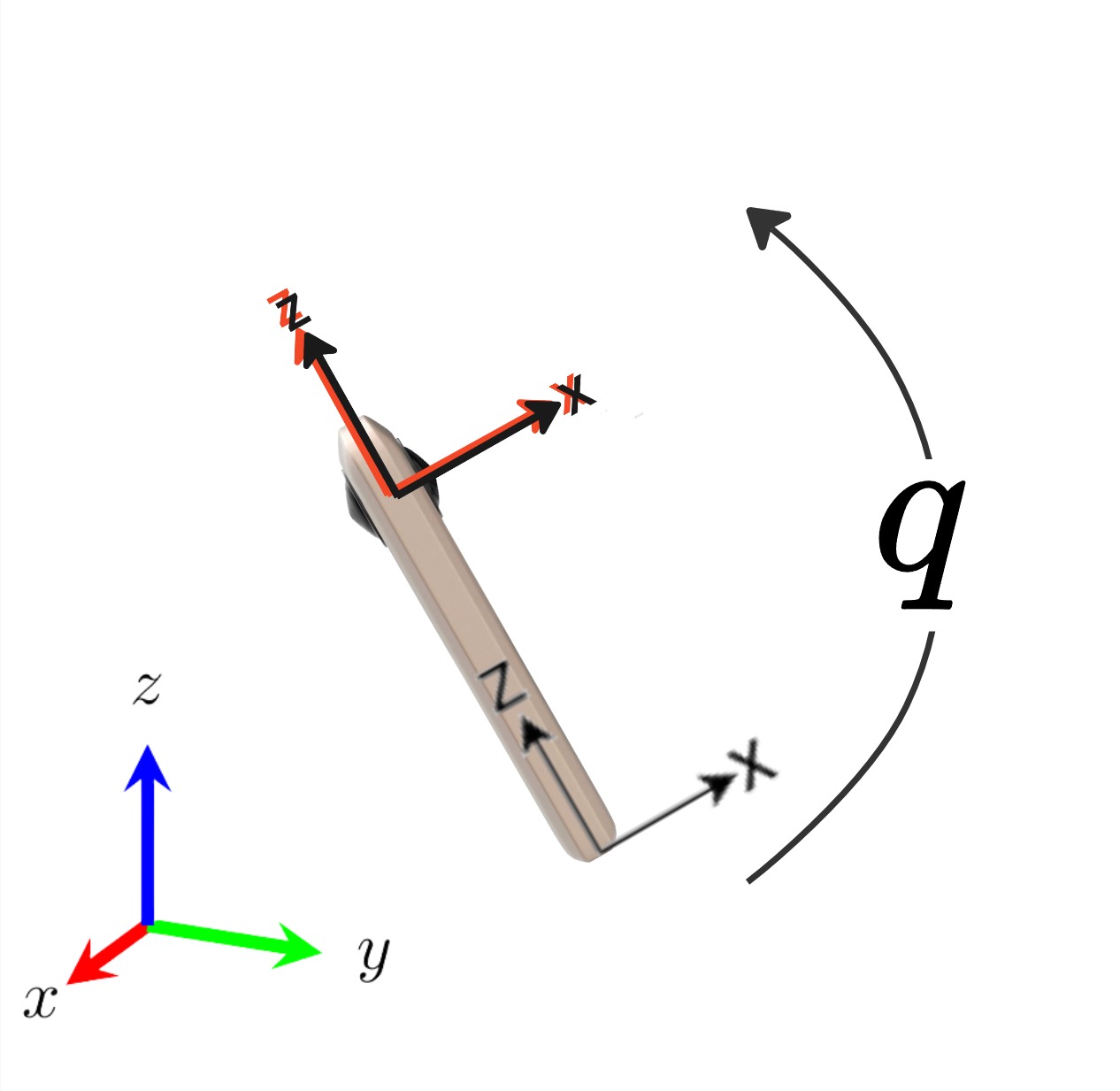}}
\subfigure[]{\label{fig:Cam_corr}\includegraphics[width=.3\columnwidth]{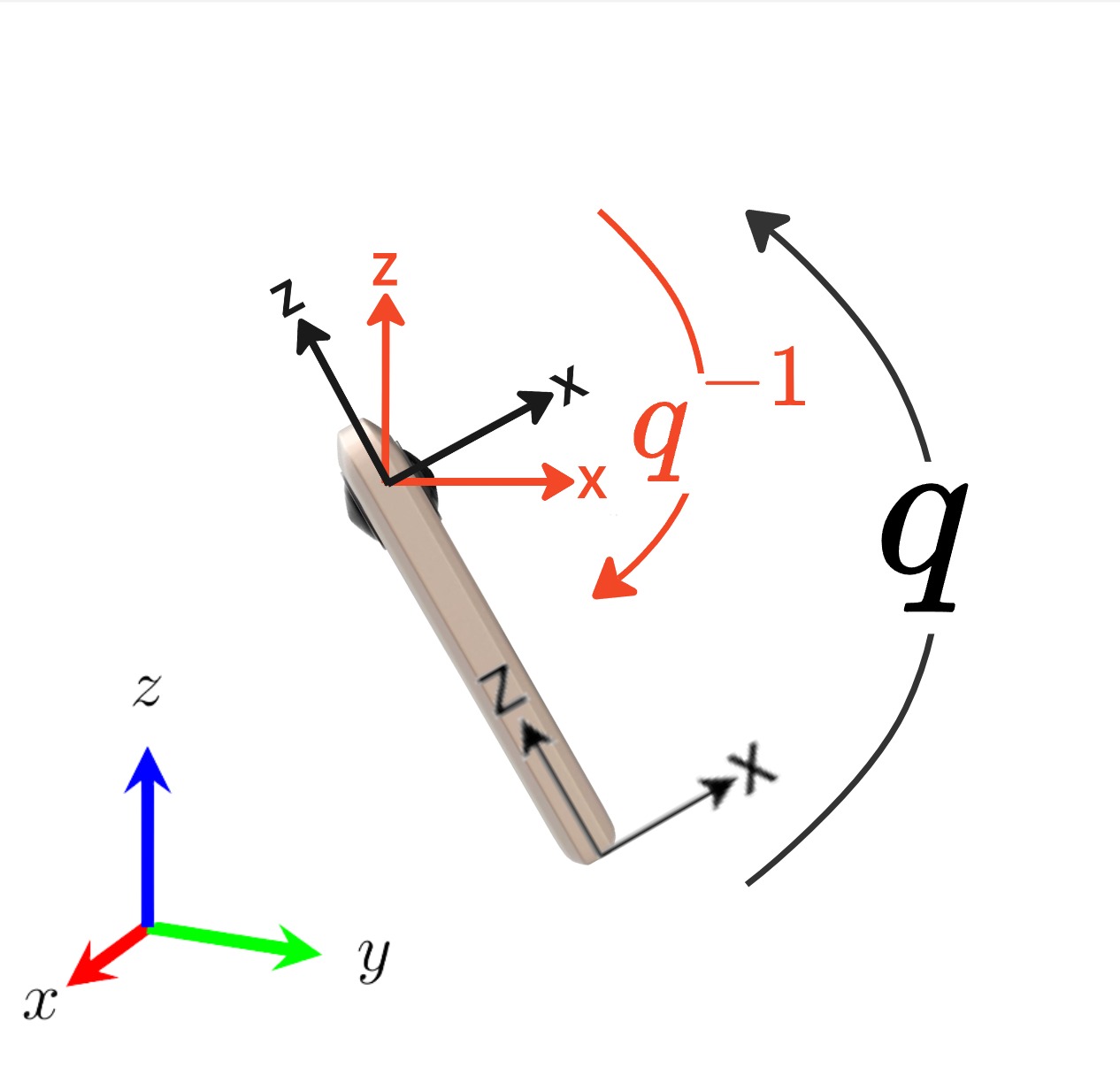}}
\caption{Unwinding the camera rotations by rotating a virtual camera frame (unwound camera frame).}
\label{fig:decouple}
\vspace{-1em}
\end{figure}

\rev{To explain the functioning of the method,}
consider the following three \rev{reference frames} 
(see, Fig.~\ref{fig:Cam_base}): the \emph{world frame} is fixed, the \emph{camera frame} is attached to the camera and moves with respect to the world frame due to the robot motion, the \emph{unwound camera frame} refers to the frame with respect to which the user viewpoint is defined. The camera frame and the unwound camera frame are coupled in the sense that by default, any rotation that affects the camera frame affects the unwound camera frame as well. 
Initially, all viewpoint references have the same orientation; that is, they can be overlayed by just applying a translation. As the camera moves, the orientation of the camera frame with respect to the world frame changes. Suppose the camera frame is rotated by $q$, in which $q$ is the unit quaternion encoding the relative orientation of the camera frame with respect to the world frame (Fig.~\ref{fig:Cam_rot}). If the unwound camera orientation with respect to the camera frame is not modified, any rotation that the camera undergoes will be experienced by the users as well, even if they do not move their heads. In order to avoid this situation, the unwound camera frame is rotated by $q^{-1}$, that is the inverse of $q$ so that the orientation of the unwound camera frame is matched with the world frame at all times (Fig.~\ref{fig:Cam_corr}). \rev{This operation ensures that the user does not experience any rotations originating from the camera rotations (see Fig.~\ref{fig:stabilViewpoint}) as opposed to viewing the rotations (see Fig.~\ref{fig:unmodViewpoint}), while preserving the user's ability control the viewpoint through head movements.}
In the following, we make this 
explanation precise and describe how it is implemented in practice. 

Let $q_{C, k}$ and  $q_{U,k}$ be two unit quaternions corresponding to the orientation of the camera frame and the unwound camera frame with respect to the world frame at time step $k$, respectively. 
Recall that if no modifications are applied $q_{U,k}=q_{C, k}$. To prevent the user from experiencing robot rotations, at each time step $k$ we set $q_{U,k}$ as
\begin{equation}\label{eq:UnvindingEq}
	q_{U,k} := q_{U,k} \cdot q^{-1}_{C,k}, 
\end{equation}
in which, $\cdot$ is the operator denoting quaternion multiplication. We used a complementary filter \cite{Mahony} to estimate the camera orientation at time $k$, that is, $q^{-1}_{C,k}$. In the ideal case of no error in the camera orientation estimate, the assignment defined in Eq.~\eqref{eq:UnvindingEq} ensures for all $k=0,1,\dots$ that $q_{U,k} = (1 , 0 ,0 , 0)$, that is, the identity quaternion corresponding to no rotation with respect to the world frame. Therefore, the only rotations experienced by the users will be caused by their head motions only. In practice, the error between the estimated value and the ground truth, as measured by the metric proposed in \cite{MetricsQuaternion}, increases over time. However, in our experiments we found the error to remain sufficiently low (on average it reached to $0.14$ radians after 2 minutes of operation) making it unnoticeable to the human eye, therefore, making the method suitable for real-time applications.

Let $\mathbb{H}$ be the set of all unit quaternions, corresponding to the set of all rotations about the origin of $\mathbb{R}^3$. In this work, unlike previous studies which considered rotations and translations in a two-dimensional plane \cite{Unwinding, cash2019improving}, we consider a camera moved by a 6 degree-of-freedom robot arm. Therefore, the set of camera rotations is the set $\mathbb{H}$. The camera used in this study is a \threesixty camera which allows to unwind any rotation within $\mathbb{H}$, satisfying the condition identified in \cite{Unwinding}.

\section{Hypotheses}
To evaluate the application of the unwinding rotations method in immersive robotic telepresence with a physical robot and a \threesixty~camera, we designed an experiment in which the camera mounted on a robotic arm moved through the RE. Study participants experienced pre-recorded videos through an HMD under two conditions, \emph{unwound rotations} (UR) and \emph{coupled rotations} (CR).
Participants were asked to complete primary and secondary tasks, and their performance and VR sickness post-exposure were compared. We tested the following directional hypotheses:

\begin{enumerate}[label=\textbf{H\arabic*}]

\item VR sickness will be lower in the UR condition. \label{hyp:sickness}
\item 
Participants will have better performance in the secondary task in the UR condition. \label{hyp:task}

\item 
Participants will prefer the UR condition. \label{hyp:preference}

\item Participants will find the UR condition more comfortable. \label{hyp:comfort}

\end{enumerate}

The hypotheses are based on the findings of previous studies on the unwinding rotations method \cite{Unwinding, cash2019improving}, as well as previous work showing that VR sickness can impair task performance \cite{attention,Wu_Zhou_Li_Kong_Xiao_2020}. 

\section{Methods}
All protocols and procedures were approved by the local ethical review board. 
The study hypotheses, sample size and analyses were preregistered\footnote{ \url{https://osf.io/x9sfh/?view_only=a749d49ca14c458d997bffc2ca50fe18}}.

\rev{\subsection{Apparatus}}\label{sec:user_study}
Fig.~\ref{fig:setup} shows the setup used in this study.
During the experiment, the participants sat on a swivel chair without wheels, allowing them to rotate freely without any translation. They watched a prerecorded video stream of the RE corresponding to either CR or UR conditions. 
For a visual comparison between these two conditions, please refer to the supplemental video: Coupled Rotations and Unwound Rotations comparison.mp4.
The RE was designed to resemble an office environment containing office furniture, storage boxes, and panels to occlude parts of the environment from particular viewing points. At the center of the room was a 
six degree-of-freedom robot arm 
equipped with a $360^\circ$~camera attached to its end-effector (see Fig.~\ref{fig:setup}). Additionally, there were objects (see Fig.~\ref{fig:SwitchBox}) and alien stickers (see Fig.~\ref{fig:BonusTask}) related, respectively, to the primary and secondary tasks that the participants were asked to perform. 

The robot arm moved within the room twice, following a predefined trajectory, and two videos were recorded. Each video lasted exactly 4 minutes and 40 seconds, a duration which, based on the pilot study, was long enough to induce VR sickness symptoms but did not tend to result in premature dropouts from the experiment. 
In both cases, the robot followed the same trajectory. To mitigate the learning effect of being exposed to an identical environment twice, two variants of the RE were prepared. These two environments were essentially the same except that different task-related objects appeared in different places in the RE. We did not wish to allow participants to control the movements of the robot because this would introduce unique path trajectories based on individual choices that might induce more or less sickness and confer better or worse chances at task success, thereby confounding the effect of decoupled rotations. We furthermore opted to use a pre-recorded video instead of a live-stream video to avoid confounding the study with arbitrary delays and artifacts, ensuring that the experience was exactly the same for each participant, though the same type of experiment could be run live in principle (see \S\ref{sec:UnwindingRotations}). 

The RE was viewed by the participants through an HMD. To achieve this, a virtual environment (VE) was designed in Unity which contains a sphere on which the 360° video stream of the RE was projected. 
Because the camera output is the equirectangular projection of a 360° image, 
video frames can be mapped onto a sphere without introducing distortions.
This VE can be observed through a Unity XR rig object which contains a virtual camera in the center of the sphere. Frames captured by this camera were displayed on the HMD screen. In such a setting, the user controls the orientation of the virtual camera, hence he or she is allowed to freely look in any direction within the RE.

The robot path was designed so that the robot end-effector carrying the camera passed through six waypoints.
The waypoints were selected mainly in spots that were obstructed at the beginning of the experience; for example, the other side of the hanging panel, visible on the right side of Fig.~\ref{fig:setup}. 
The robot followed the path with a constant speed, making a five second pause at each of the six waypoints. The waypoints and the respective pauses correspond to places in the RE where the objects related to the primary task were placed (see \S\ref{sec:measures} for a description of the primary task). Each video contained only three of the objects related to the primary task, whereas the robot paused at each of the waypoints for each video.
We made sure that the orientation of the camera changed between each waypoint; for example, at one waypoint the orientation was opposite to the initial orientation (i.e., upside down relative to the starting configuration). 
The robot followed the path at a constant speed of $100.0~mm/s$ and the maximum magnitude of linear acceleration/deceleration was $50.0~mm/s^2$.

The videos were recorded with a RICOH THETA V \threesixty~camera in 3840×1920 resolution and a frame rate of 29.97 fps. An Oculus Quest 2 was used as the HMD, with a resolution of 1832 × 1920 per eye and a refresh rate of 80 Hz, connected to a PC with Unity through Wi-fi with Meta Quest Air Link.
The WIT Motion WT931 IMU was attached to the camera base and used during the video recordings to collect the data needed to estimate the orientation of the camera. 

\vspace{2mm}
\subsection{Procedure}

Upon arrival, participants were asked whether they were experiencing a headache or nausea, and if so, asked to reschedule for another day. Before the beginning of each session, each participant signed an informed consent form and filled out a demographic questionnaire. 
Next, a high-level study description of the task was read to the participant. 
After that, the participant was asked to read the task description.
The participants were then asked to sit on a swivel chair. 
Directly before the exposure to the RE, short instructions were given on how to adjust the headset and use a controller to complete the task.  
After the participant confirmed readiness, one of the two 360° videos corresponding to one of the conditions (UR or CR) was played through the HMD. 
The videos were presented in a counterbalanced sequence, ensuring that each video and condition pair was viewed as the first and second an equal number of times by both men and women.
After each session, participants were asked to answer questionnaires, starting with the Simulation Sickness Questionnaire (SSQ) \cite{SSQ}.
After the second session, additional questionnaire items gauged users' preferences between the two experiences. 
The second session was scheduled on a different day to prevent potential carryover effects caused by an accumulation of VR sickness symptoms over sequential sessions \cite{Duzmanska}. Participants saw one of four counterbalanced sequences based on the condition and RE variant.

\begin{figure}[t]
\centering   
\subfigure[Nominal state]{
   \includegraphics[width=.45\columnwidth]{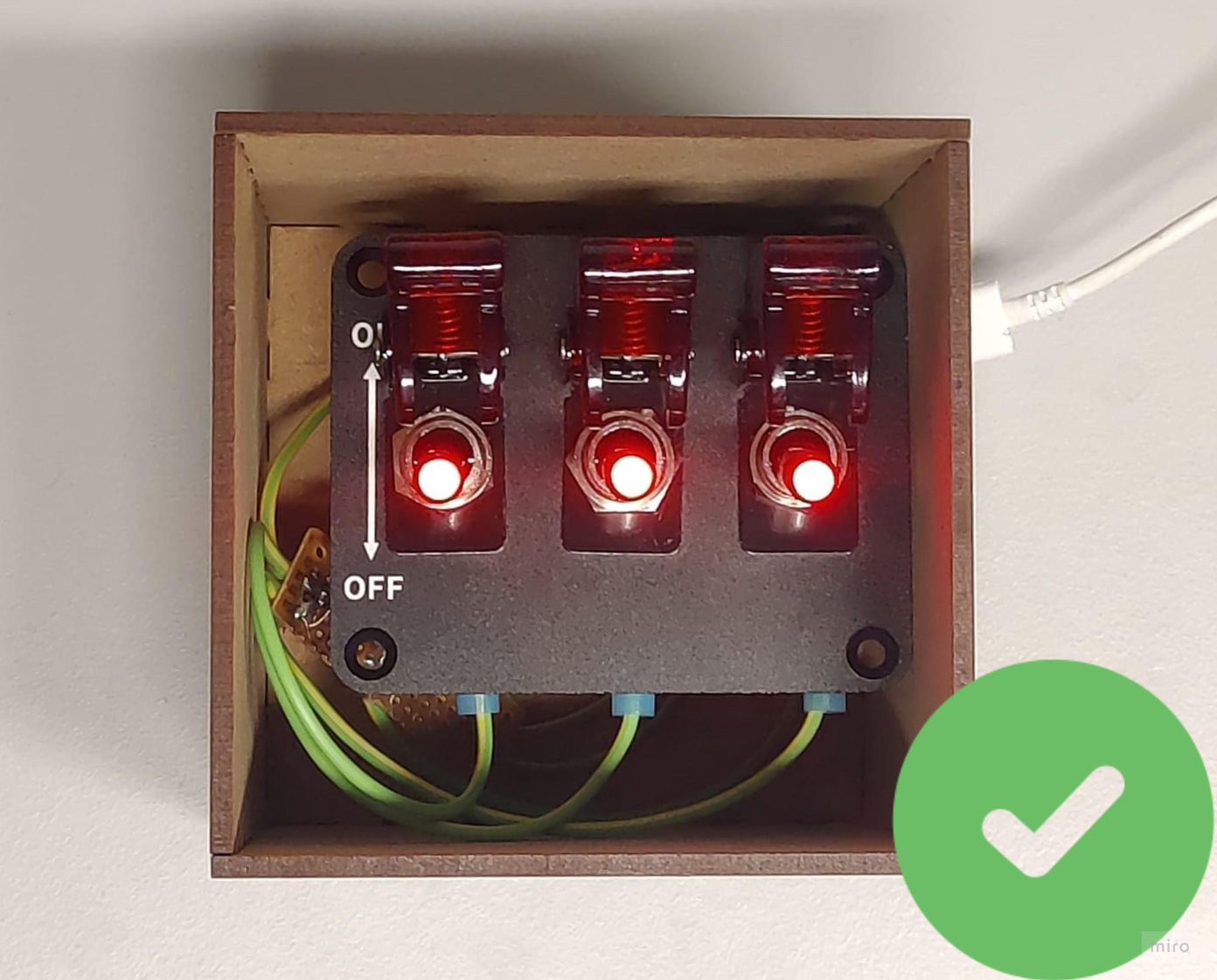}
     \label{fig:SwitchBoxCorrect}}
\subfigure[Malfunctioning state]{
    \includegraphics[width=.45\columnwidth]{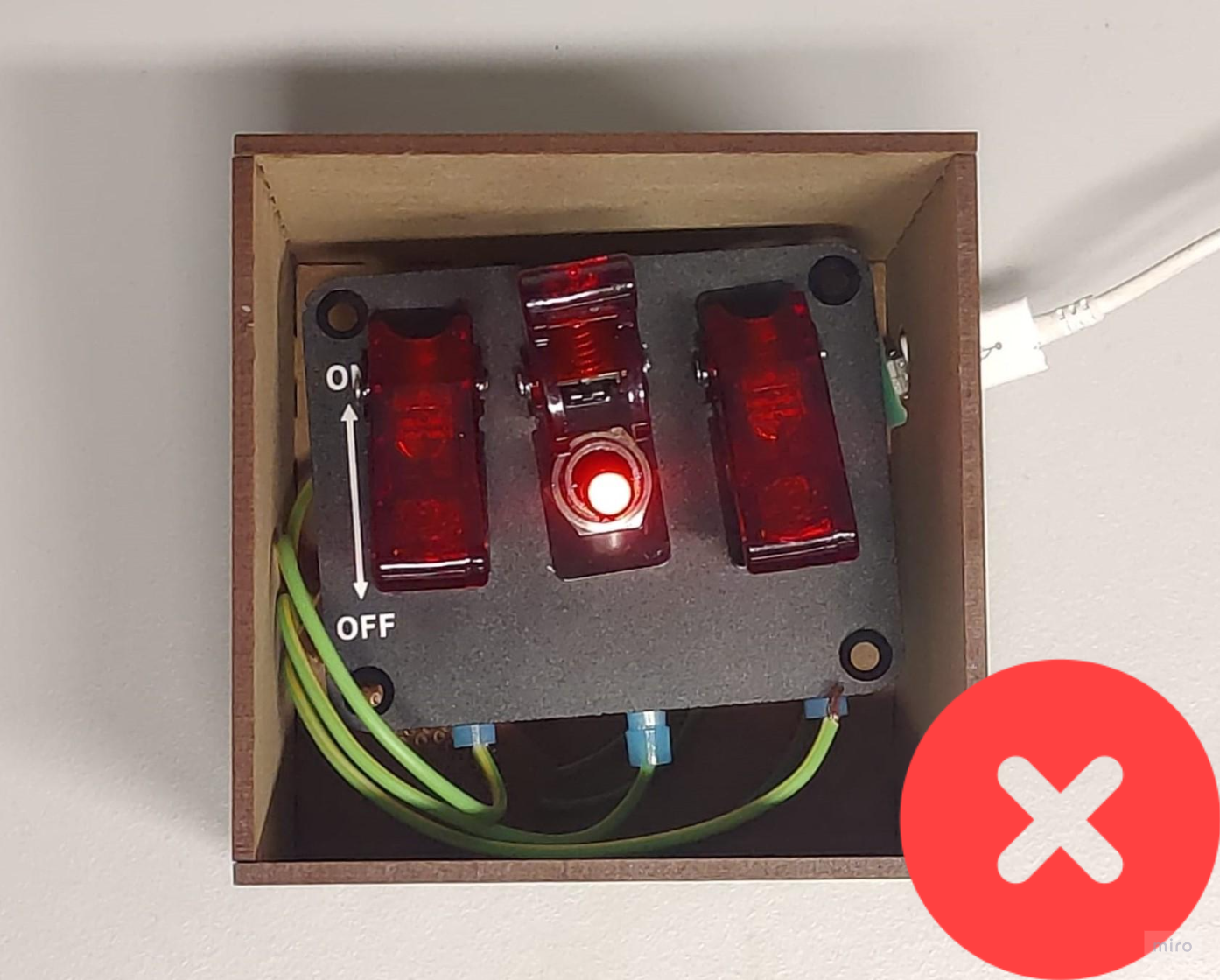}
    \label{fig:SwitchBoxWrong}
}
\caption{An example object (a switch box) used for the primary task shown in two possible states.
}
\label{fig:SwitchBox}
\vspace{-1.5em}
\end{figure}
\vspace{-4mm}

\rev{\subsection{Design}}\label{sec:measures}
To measure the participants' performance in paying attention to the anomalies in the environment, a primary 
and a secondary task 
were developed. \rev{These tasks required the participants to look around to inspect a remote environment, which is a representative use case for telepresence robots~\cite{Su_Chen_Zhou_Pearson_Pretty_Chase_2023}.
}
The primary task was to detect and determine the condition of the devices placed at the waypoints along the robot path. \rev{Even if the robot stopped at these waypoints, the users might still have needed to rotate to see the device.} Each experience contained three objects for which the participant needed to press a button on the controller indicating whether the object was malfunctioning. Fig.~\ref{fig:SwitchBox} illustrates an example object in two possible states. The primary task was intended to be relatively easy, and no differences between conditions were predicted.

\begin{figure}
\centering   
\subfigure[]{
    \includegraphics[width=.45\columnwidth]{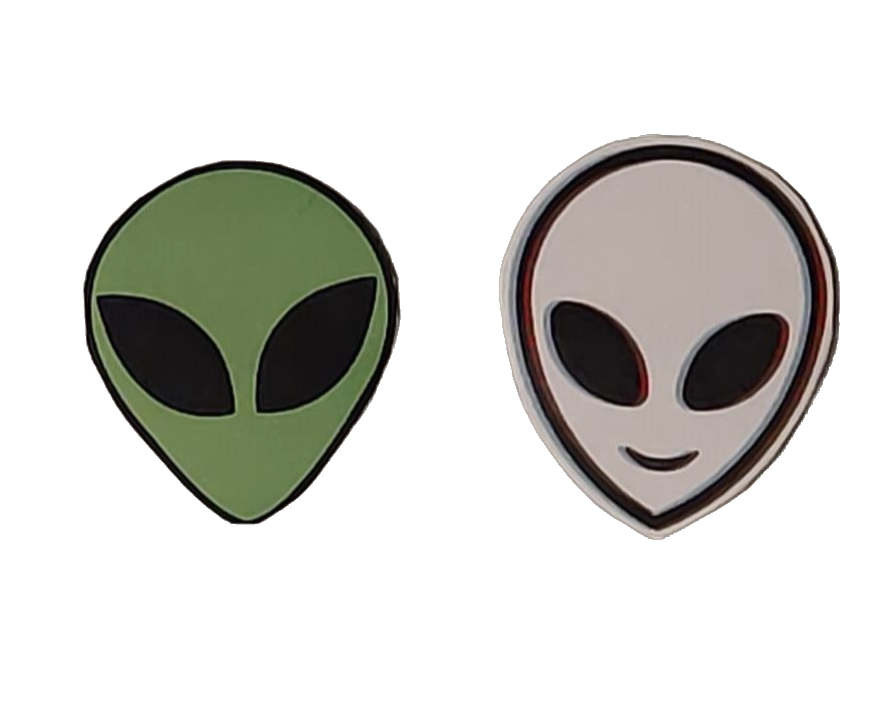}
    \label{fig:AlienOptions}
}
\subfigure[]{
    \includegraphics[width=.45\columnwidth]{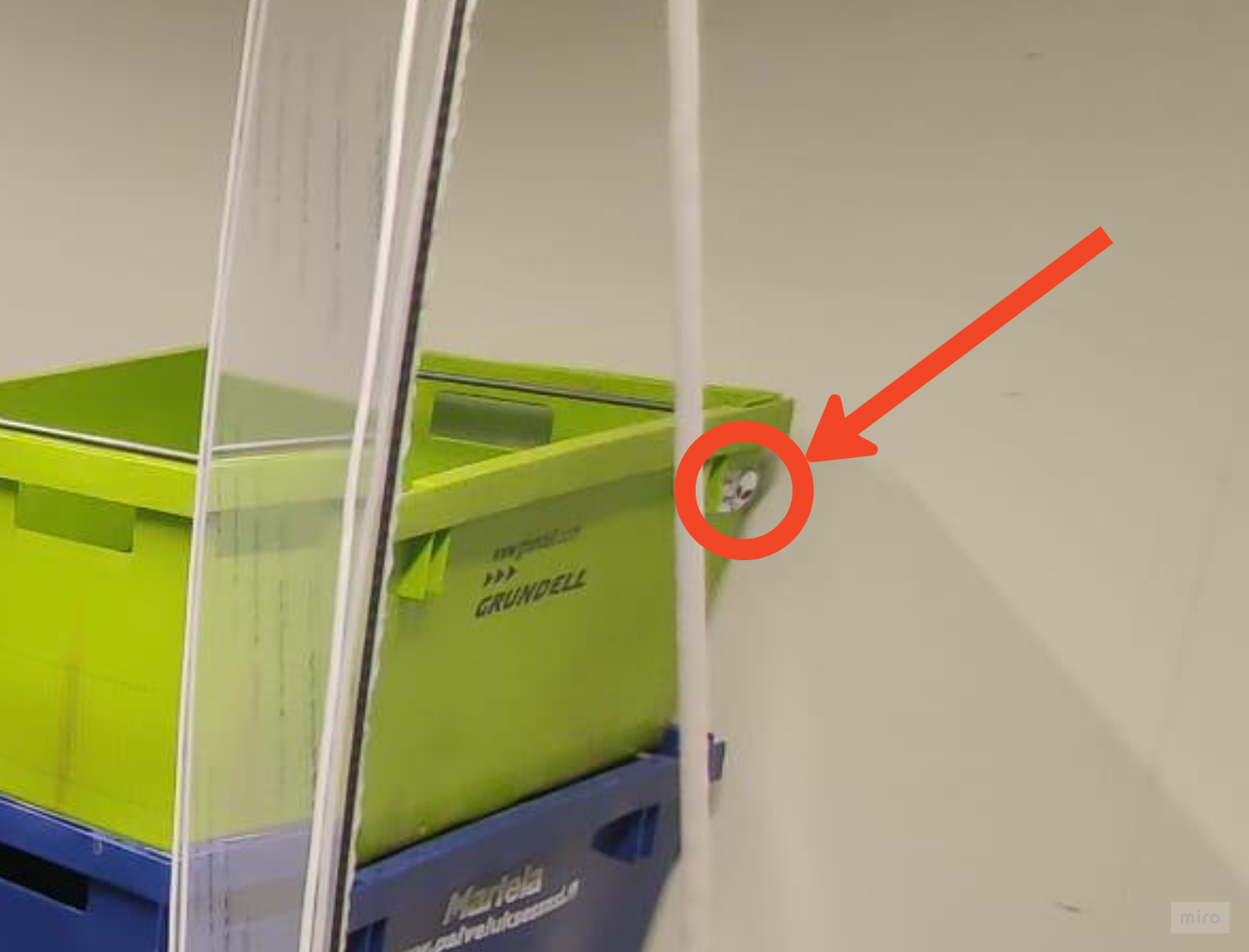}
    \label{fig:AlienRoom}
}
\caption{Secondary task of detecting and tagging aliens in the environment. (a) The two types of alien stickers used in the experiment. (b) An instance of alien placement in the environment.}
\label{fig:BonusTask}
\vspace{-1em}
\end{figure}

We included a secondary task to strain participants' attentional resources and measure whether their ability to pay attention to more subtle aspects of the environment might differ as a function of experiencing coupled or unwinding rotations. We aimed to simulate a realistic scenario relevant to robotic telepresence in which the remote operator also needs to notice possible abnormalities while doing routine inspection of predefined features.
The secondary task was to find and mark/shoot \emph{aliens} scattered in the environment. To this end, ten alien head stickers were attached to different elements in the room in less visible places.
The alien stickers had two color versions, as presented in Fig.~\ref{fig:AlienOptions}. An example of alien placement in RE is visible in Fig.~\ref{fig:AlienRoom} (zoom of Fig.~\ref{fig:setup}). \rev{Even though each alien sticker was visible along the robot trajectory for some time, the users needed to look in the correct direction to see it. Therefore, they needed to actively look around to spot the alien stickers.}
Marking the aliens was accomplished by clicking predesignated buttons on the VR controller.
A demonstration of successful completion of one of the primary objectives and two secondary objectives can be seen in the supplemental video: Primary and Secondary tasks – Participant view (Unwound Rotations).mp4.

During the experiment, we collected data by logging the timestamp and the orientation of the controller whenever a controller button was clicked. 
Additionally, when an alien marking button was clicked, a screenshot of the user screen was taken to allow later verification of whether the user correctly aimed at aliens and to eliminate duplicate shots at the same alien. 
The performance score of the primary task was the number of correctly marked devices. A device was deemed correctly marked if the participant clicked the button indicating the nominal state if the device was in the nominal state or vice versa at the correct point along the robot path. 
The performance score of the secondary task was the number of unique aliens marked. 
The scores for the secondary task were manually calculated by research assistants who were blind to the experiment's hypotheses, on the basis of the screenshot and buttons log.

After each VR session, participants filled out the simulator sickness questionnaire (SSQ) \cite{SSQ}, the NASA - Task Load Index (NASA-TLX) questionnaire \cite{NasaTLX}, and gave a comfort score rating.
In the SSQ, participants self-report 16 symptoms of sickness on a 4-point scale (none/slight/moderate/severe). 
The total weighted score based on those 16 values 
was calculated.
A higher score corresponds to greater levels of sickness experienced.
In the NASA-TLX questionnaire participants self-report six aspects of workload (mental demand, physical demand, temporal demand, performance, effort, and frustration).
\rev{The average of six values (RAW TLX) were used for further analysis.}
For the comfort score, participants rated the motion experienced during each experience on a 7-point scale (from ``Very uncomfortable" to ``Very comfortable"). 

After the second session, participants were asked to choose their preferred session and answer open-ended questions about their reason for this choice. 
To avoid bias, participants were informed about the difference between conditions only after having completed all questionnaires. Each session lasted approximately 15-20 minutes. Subjects received a custom clothing patch and a €5 gift card for participating.

\vspace{2mm}
\subsection{\rev{Pilot Study}}
\rev{
We conducted a separate pilot study with 20 participants (who were not allowed to participate in the main study) in order to test our protocol and estimate effect sizes to be used in \textit{a priori} power analysis in the main study. The pilot was identical to the main study except in that the headset was connected to the PC via a wired connection, whereas in the main study it was connected wirelessly to give participants the ability to rotate their chair more freely, and that the pilot study queried comfort using a binary forced-choice between conditions at the end of the study instead a separate Likert rating after each condition as in the main study. 
}

\rev{
The difference in SSQ scores between the UR ($Mdn = 20.57$) and CR ($Mdn = 39.27$) conditions was statistically significant, as indicated by the Wilcoxon signed rank test (two-tailed), $Z = 2.30$, $p = 0.021$, $r = 0.51$.
The difference between the number of targets detected in the secondary task, for the UR ($M = 3.3$, $Mdn = 2.0$) and CR ($M = 2.1$, $Mdn = 2.0$) conditions (max = 10) was statistically significant, as indicated by the Wilcoxon  signed rank test (two-tailed), $Z = 2.1$, $p = 0.034$, $r = 0.42$. The difference in RAW-TLX scores between the UR ($Mdn = 5.9$) and CR ($Mdn = 6.3$) conditions was not statistically significant, as indicated by the Wilcoxon signed rank test (two-tailed), $Z = 0.17$, $p = 0.86$, $r = 0.05$.
}

\rev{
When asked for a preferred method, 75\% (15 of 20) participants selected the UR condition. 
User preference was analyzed using an exact binomial test against 50\% (no preference between conditions).
A binomial test indicated that this tendency in preference is significant ($p = 0.041$, $g = 0.25$).
Furthermore, when asked for a more comfortable method, 90\% (18 of 20) participants selected the UR condition. A binomial test was performed showing that the UR condition was found significantly more comfortable ($p < 0.001$, $g = 0.40$).
}

\subsection{Participants}

We used \textit{a priori} power analysis via G*Power software \cite{faul2007g} to determine our sample size, \rev{focusing primarily on our SSQ result. Cohen's $dz$ was derived for the purposes of power analysis by dividing the mean of the differences between the two conditions by the standard deviation of those differences. Based on a pilot study ($n = 20$; described above)} we estimated that 36 participants would provide at least $95\%$ power to detect our estimated SSQ total score difference ($dz = 0.61$) \rev{under assumptions of a one-tailed Wilcoxon signed-rank test with a minimum possible average risk error (min ARE) distribution. A sample size of 36 participants was preregistered prior to data collection. Additionally, this sample size would provide at least 95\% power to detect our estimated comfort difference ($g = 0.40$), at least $90\%$ power to detect our estimated preference difference ($g = 0.25$), and at least $80\%$ power to detect our estimated secondary task performance difference ($dz = 0.50$).}

Participants were recruited from the university and the greater community and signed up for the study through an online recruitment page. Thirty-six volunteers participated. 
Of those 36 individuals, 14 were women, 20 were men, one chose ``other" gender, and one preferred not to disclose their gender. The average age of the sample was 27.6 years (range: 19-41). 
There were five participants 
who played computer games daily, nine played multiple times a week, three played once or twice a week, six played once or twice a month, five played once or twice a year, five played only a few times in their lives, and three answered that they never played any computer games.
In case of virtual reality systems, none of the participants reported using them daily, two used them several times a week, three used them once or twice a week, four used them once or twice a month, six used them once or twice a year, 16 used them only a few times in their lives, and five reported having no prior experience with Virtual Reality systems.
Among all participants, 18 wore glasses or contact lenses, and one person was color blind.

\begin{figure}[b]
\centering
\subfigure[SSQ Scores]{\label{fig:SSQ}\includegraphics[width=0.32\linewidth]{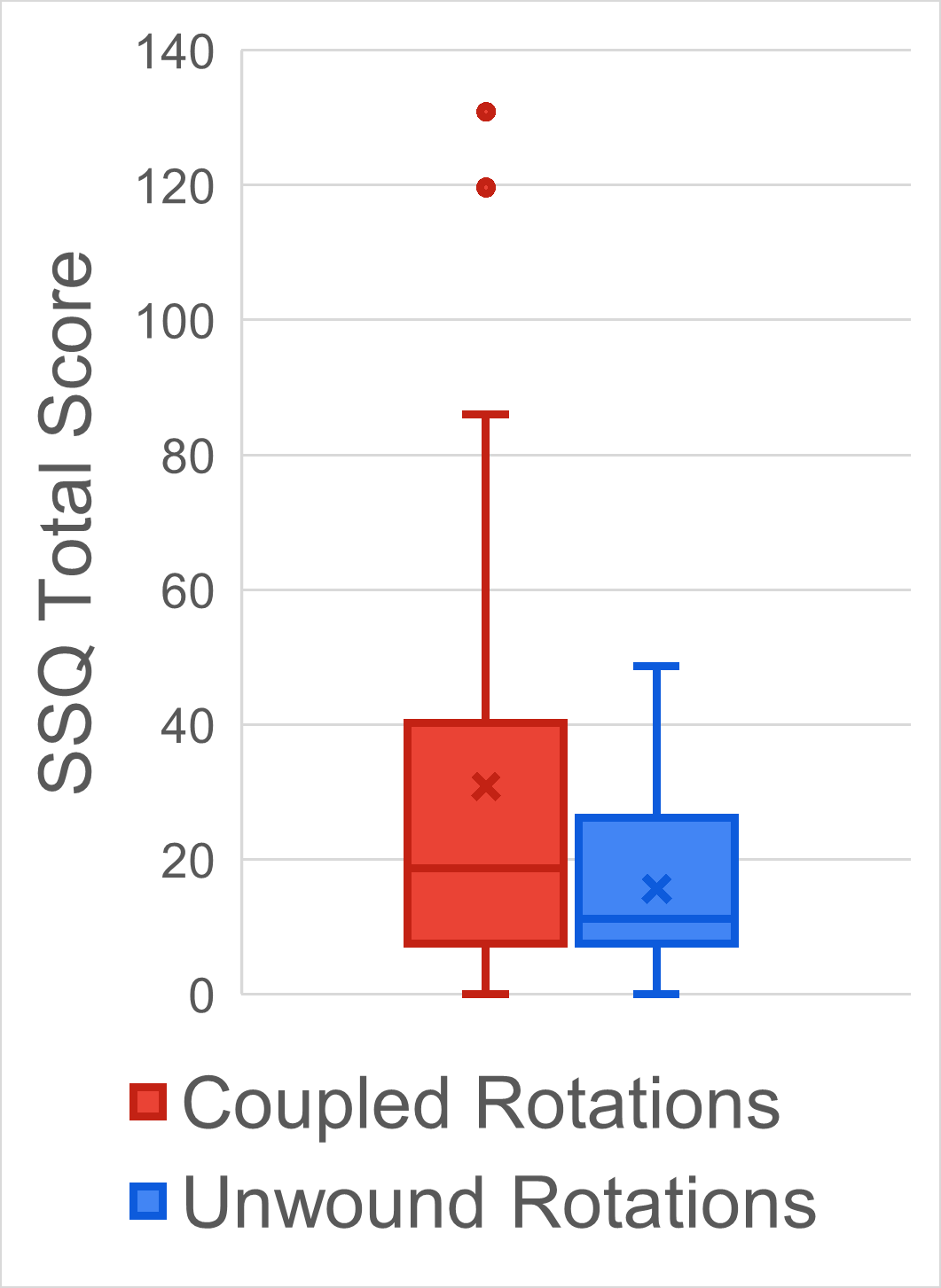}} \hfill
\subfigure[Task Score]{\label{fig:aliensScore}\includegraphics[width=0.32\linewidth]{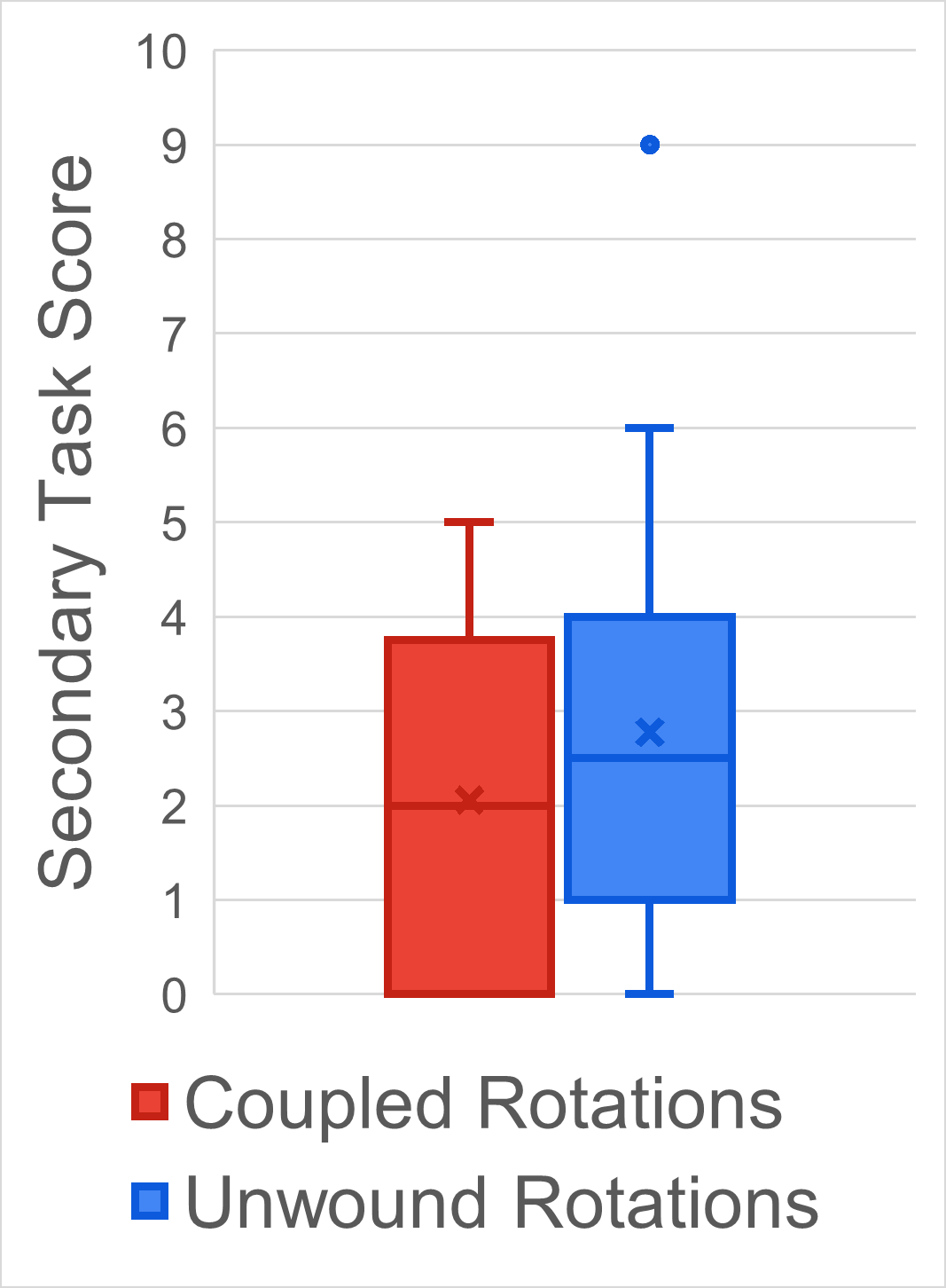}} \hfill
\subfigure[Comfort]{\label{fig:comfort_score}\includegraphics[width=0.32\linewidth]{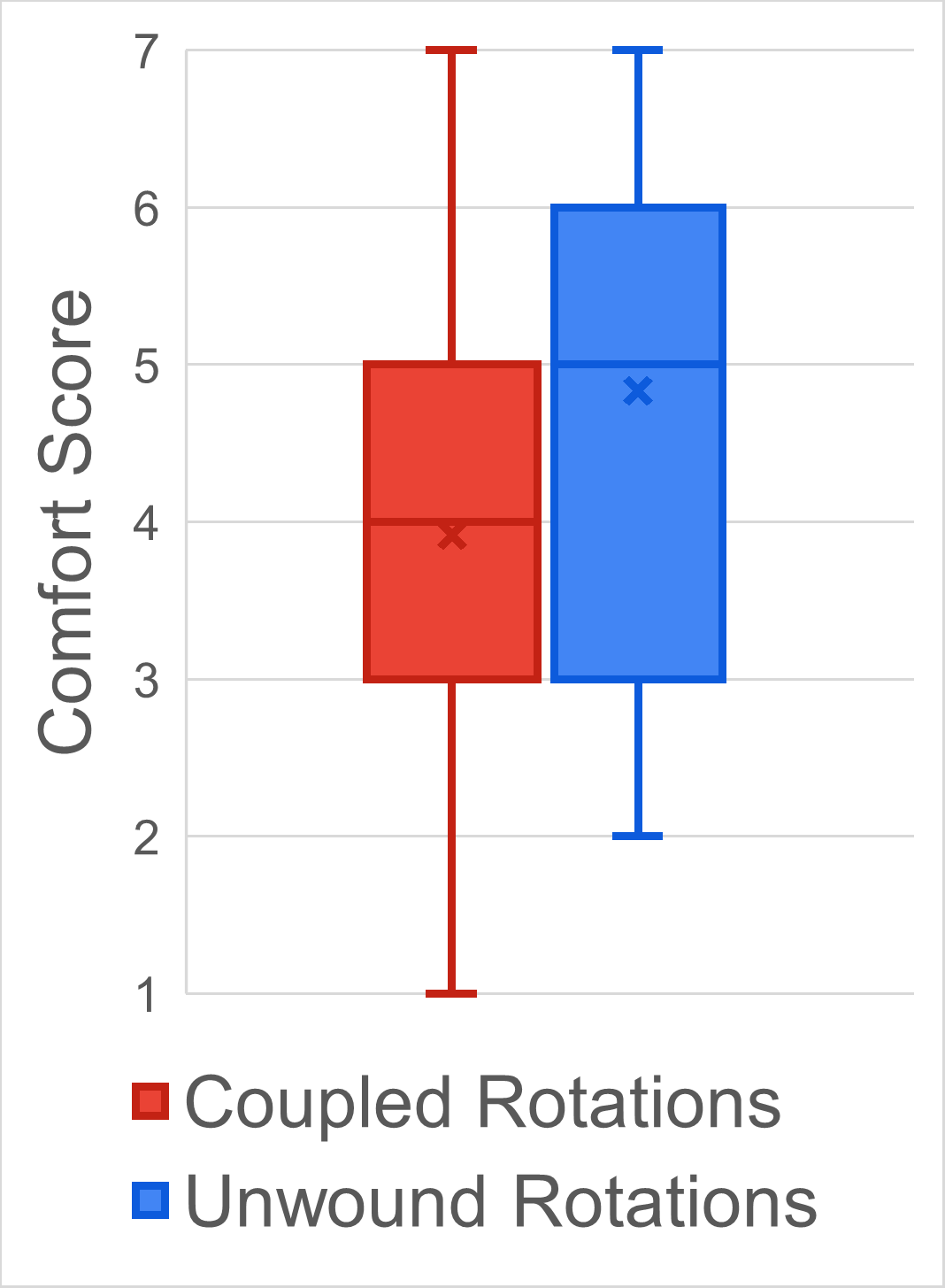}}
\caption{Comparisons of distributions corresponding to different conditions. 
The box represents the interquartile range, with whiskers extending to the minimum and maximum values excluding outliers, which are plotted as dots. A horizontal line marks the median and a cross indicates the mean.
}
\end{figure}

\section{Results}
All confirmatory tests were performed in R statistical analysis software with (one-tailed) statistical significance levels set to 0.05, and all exploratory tests were performed as a two-tailed test with the same significance level requirement. 

\subsection{Confirmatory Results}
A Wilcoxon signed-rank test was performed to compare the differences between the total weighted SSQ scores for the two conditions, UR ($Mdn = 11.22$) and CR ($Mdn = 18.70$) (see Fig.~\ref{fig:SSQ} for the score distributions).	 
The test indicated that UR resulted in significantly lower SSQ scores compared to CR,
$Z = 3.05, p = .001, r = 0.51$, supporting the hypothesis \ref{hyp:sickness}.

Wilcoxon signed-rank tests were used to compare differences in scores in the primary (two-tailed test; not preregistered) and secondary (one-tailed test; preregistered) tasks. Scores on the primary task did not differ significantly between the UR ($Mdn = 3$) and CR ($Mdn = 3$) conditions (max $=3$), $Z = 0.41, p = .68, r = .05$. This outcome was expected since we intended the primary task to be achievable by everyone regardless of the condition. 
\rev{The difference between the number of targets detected in the secondary task, for the UR ($M = 2.8$, $Mdn = 2.5$) and CR ($M = 2.1$, $Mdn = 2.0$) conditions (max $= 10$; see Fig.~\ref{fig:aliensScore} for the score distributions), fell just short of statistical significance, $Z = 1.60, p = .055, r = 0.26$,
indicating that effect of condition change on secondary task performance is inconclusive, failing to support \ref{hyp:task}.}

When asked for a preferred session, $78\%$  (28 out of 36) participants selected the UR condition.
User preference was analyzed using an exact binomial test against $50\%$ (no preference between conditions). 
The test indicated that the participants significantly preferred the UR,
$p < .001$. 
This result provides support to \ref{hyp:preference}, indicating a strong preference for the UR condition.

Comfort was analyzed using the Wilcoxon signed-rank test on the Likert ratings given to the condition UR ($Mdn = 5.0$) and CR ($Mdn = 4.0$; see Fig.~\ref{fig:comfort_score} for the score distributions).
The test indicated that UR resulted in a significantly higher comfort score compared to CR,  
$Z = 2.19, p = .014, r = 0.36$.
This finding supports \ref{hyp:comfort}, suggesting that the UR condition was more comfortable to participants.

\subsection{Exploratory Results}
\rev{Perceived workload was measured using the NASA-TLX. The RAW TLX value (average of all subscales \cite{NasaTLX}) and subscales medians are given in Table \ref{tab:nasa}.  
When comparing the NASA-TLX RAW scores with a Wilcoxon's signed rank test, we did not observe statistically significant differences between the values of the UR ($Mdn = 7.9$) and CR ($Mdn = 7.6$) conditions ($Z = 0.78$, $p = 0.44$, $r = 0.012$).}

\begin{table}[h]
\centering
\begin{tabular}{|l|ll
|}
\hline
\multirow{2}{*}{\textbf{TLX subscale}} & \multicolumn{2}{c|}{Median}
\\ \cline{2-3}
                              & \multicolumn{1}{l|}{UR} & CR
\\ \hline
Mental demand                 & \multicolumn{1}{l|}{9.5} & 9.0
\\ 
Physical demand               & \multicolumn{1}{l|}{4.5} & 3.5
\\ 
Temporal demand               & \multicolumn{1}{l|}{7.0} & 3.5
\\ 
Performance                   & \multicolumn{1}{l|}{7.5} & 9.0
\\ 
Effort                        & \multicolumn{1}{l|}{10.0} & 8.5
\\ 
Frustration level             & \multicolumn{1}{l|}{3.0} & 5.0
\\ \hline
\rev{\textbf{RAW TLX}}             & \multicolumn{1}{l|}{7.9} & 7.6
\\ \hline
\end{tabular}

\vspace{2mm}
\caption{\label{TLX_table} \rev{Medians
of NASA-TLX subscales and RAW TLX across UR and CR conditions.}}
\label{tab:nasa}
\vspace{-1em}
\end{table}

\begin{figure}[thpb]
     \centering     \includegraphics[width=\linewidth]{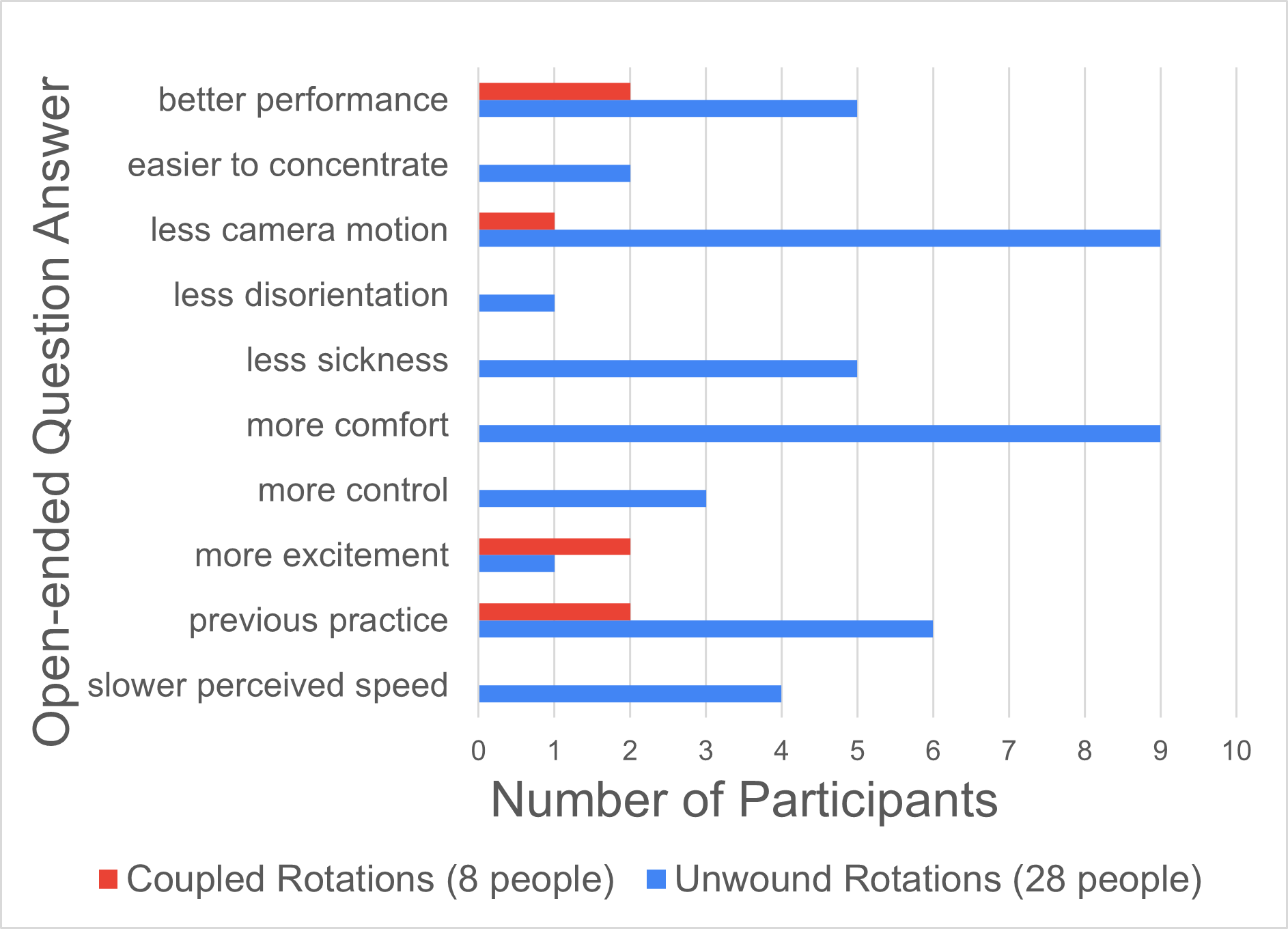}
     \vspace{-2em}
     \caption{Bar chart depicting instance counts of individual reasons given for participants' preference of one condition over another.} 
     \label{fig:prefered}
\end{figure} 

To better understand the reasons governing the participants' preference of the two different conditions, we analyzed their responses to the open-ended question asking the reason for their preference.
Users' justifications for their preference were coded by two researchers independently, which were then harmonized into consistent themes and counted. 
Figure \ref{fig:prefered} shows the frequently found codes from open-ended questions regarding the reasoning behind the preferred method. 

The reasons why the participants chose the UR condition as the preferred one matched our expectations.
Several people directly noticed that their symptoms of VR sickness were lower after the UR condition (\textit{``The second session didn't make me as motionsick as the first one."})
Although a handful of participants selected the CR condition, the reasoning behind this choice was often based on the enjoyment aspect.
Additionally, the ``better performance" aspect was mentioned more frequently in relation to the UR condition.

\section{Discussion}
The objective of this study was to utilize the unwinding rotations method in a nonsimulated setting to test its effectiveness in reducing VR sickness (\ref{hyp:sickness}) and improving task performance (\ref{hyp:task}).
Moreover, based on previous research, we expected that the unwound rotations condition would be preferred by users (\ref{hyp:preference}) and rated as more comfortable (\ref{hyp:comfort}). 
The results of the study supported hypotheses \ref{hyp:sickness}, \ref{hyp:preference}, and \ref{hyp:comfort}. \rev{However, \ref{hyp:task} did not replicate the pilot result.}

Our hypothesis related to reduced VR sickness in the UR condition, that is, \ref{hyp:sickness}, was based on the results of the previous studies \cite{Unwinding, cash2019improving} that this work attempted to extend, as well as the evidence in the literature on rotational motion having a greater impact on VR sickness compared to translational motion \cite{hu1999systematic,kemeny2017,CybersicnessKemeny2020}. Participants reported significantly less intense VR sickness symptoms after they experienced the UR condition than after they experienced the CR condition. 
When the view of the user was decoupled from the motion of the camera, participants were able to determine when and in what direction they would rotate to view the environment. Thus, there was less sensory conflict between expectations and sensations when visual cues and vestibular stimulation matched as a result of physically turning the head in the UR compared to the CR condition. Furthermore, since the participants knew when rotations would occur, there were fewer unexpected visual motion cues, which is another factor that can have a significant effect in how much VR sickness is experienced \cite{Teixeira_Miellet_Palmisano_2022,Teixeira_Miellet_Palmisano_2024}. 

Even though the hypotheses \ref{hyp:preference} and \ref{hyp:comfort} were also based on the findings of the previous work, they were originally tested under different conditions considering a simulated environment and a robot capable of rotations only in a two dimensional space. 
In particular, the main concern of the authors in \cite{Unwinding} was whether the users found rotating in the chair uncomfortable or tiresome, and whether the reduction in VR sickness was sufficient to choose one condition over the other.
Our study differed from previous work in that we used a video stream from a real camera which moved in a three dimensional space. Therefore, it required the participants to execute a larger range of motion to compensate for the rotations that were unwound if they wanted to align themselves with the direction that the camera was moving.
However, participants still found the UR condition more comfortable. \rev{In constrast, exploratory analyses indicated that the UR condition elicited median increases across several dimensions of the NASA-TLX, with only the frustration and performance dimensions seeing any potential signs of improvement}.  
Similar to the previous study, main reasons for people's preference for the UR condition were related to reduced VR sickness and increased comfort, indicating that VR sickness is an important aspect in terms of usability of immersive telepresence systems.

Interestingly, 5 out of 28 participants who preferred the UR condition mentioned better performance as one of the reasons for their choice (for example, ``\emph{Motion is less in the method two and it was easy to identify objects and concentrate than in method one.}'', in which method two refers to the UR condition). However, we did not observe a reliable quantitative difference in their performance across conditions. \rev{Indeed, we failed to reject the null hypothesis with respect to hypothesis \ref{hyp:task}, which stated that participants would have better task performance under the UR condition.}
The lack of \rev{an observed statistical} difference in secondary task performance may be partially attributed to the specific nature of the secondary task, which required participants to actively look for changes in an unfamiliar environment. In contrast, in some potential use cases, such as an inspection task, the environment may be more familiar and predictable to the user. 
Furthermore, we expected to have the lowest statistical power to detect this effect among all hypotheses based on pilot data, \rev{in spite of the statistically significant effect detected in the pilot.} The observed trends in task performance aligned with the expected direction, perhaps indicating that the study was still underpowered to detect this effect and that a larger sample size could have yielded statistically significant differences in task performance.

Effective utilization of VR for robotic telepresence requires further investigation into methods to improve the comfort and intuitiveness of navigation in a RE.
We believe that the use of the unwinding rotations method is well-suited to inspection tasks, in which the user must maintain focus on a target regardless of the rotation of the robot platform. This method might be most helpful in situations where there are fewer constraints on robot motions compared to a mobile ground robot, such as in the case that the \threesixty~camera is mounted on a drone, submersible, or spacecraft. In these cases, the increase in the amount of camera motion may also lead to greater levels of VR sickness. 
Although we did not find a difference in task scores here, VR sickness has been shown to impair task performance \cite{attention,Wu_Zhou_Li_Kong_Xiao_2020}, so any methods reduce discomfort that is more intense than that which occurred in our study may significantly improve the user's ability to perform in the RE. 
Additionally, in immersive robotic telepresence tasks where it is important that the user is able to develop or maintain awareness of the location of objects in the surrounding environment, having control over what area is viewed at any moment in time may facilitate spatial awareness and construction of mental maps of the RE.

\subsection{Limitations and Future Work}

The secondary task had a game-like nature with small anomalies appearing for a short period of time in an otherwise static environment. 
We wanted a secondary task that would divide the user's attention and test their ability to notice small details, though perhaps this implementation resulted in a task that was too game-like relative to real-life use cases. 
The difficulty of the secondary task was set at a relatively high level, creating a near floor effect in the scores ($M = 2.42$, $Mdn = 2$, out of 10), potentially obscuring our ability to observe differences in task performance between conditions.
Thus, future work could employ the unwinding rotations method in a variety of tasks, robot types, and environments to establish more robust generalizations about the effectiveness of the method. 

We initially implemented the unwinding rotations method to work in real-time video streaming. However, we chose not to use this option in this study due to the possibility for delays and missing frames that could happen in a real-time application, which could have confounded the results. 
\rev{Thus, we used a predefined robot trajectory and prerecorded videos as this ensured that all participants experienced similar
visual stimuli.} 
In future work, extending the study setup to consider real-time video streaming \rev{with the user able to control the robot arm} would allow us to consider changing environments that the users can interact with, 
with the additional consideration that sickness and performance are likely to be significantly impacted by whatever amount of network delay is present. 
\rev{Note also that even in real-world applications, it is possible that the user would not have full autonomy over the robot's trajectory due to shared control mechanisms (see for example, \cite{selvaggio2021shared,luo2019teleoperation}). These mechanisms result in semi-autonomous robot motion, improving task accomplishment or collision avoidance that might be hampered by the delays in feedback or in control.}

\section{Conclusion}
We show that the use of the unwinding rotations method decreases VR sickness, increases comfort, and is preferred by users
when performing the inspection task in VR in an autonomously moving telepresence robot.
Despite a trend in the predicted direction, we \rev{could not statistically confirm differences in task performance between the unwound rotations and coupled rotation conditions as observed in the pilot data.} 
Thus, the unwinding rotations method shows great potential to make tasks requiring a high mental workload while moving around RE (e.g., inspection tasks) more comfortable, with, at worst, no significant detriment to task performance. 

\acknowledgments{This work was supported by the European Research Council (project ILLUSIVE 101020977) and the Academy of Finland (projects BANG! 363637, CHiMP 342556, PERCEPT 322637).}

\bibliographystyle{abbrv-doi}

\end{document}